\documentclass{article}

\usepackage{microtype}
\usepackage{graphicx}
\usepackage{subfigure}
\usepackage{booktabs} 

\usepackage{hyperref}

\usepackage[accepted]{icml2023}

\usepackage{amsmath}
\usepackage{amssymb}
\usepackage{mathtools}
\usepackage{amsthm}

\usepackage[capitalize,noabbrev]{cleveref}

\theoremstyle{plain}

\theoremstyle{definition}

\theoremstyle{remark}

\usepackage[textsize=tiny]{todonotes}

\usepackage{amsmath,amsfonts}
\usepackage{booktabs,arydshln}
\usepackage{multirow}
\usepackage{ulem}
\usepackage{pifont}
\usepackage{relsize}
\usepackage{enumitem}
\usepackage{wrapfig}
\usepackage{graphicx}
\usepackage[labelfont=bf, font=small]{caption}
\usepackage{subfigure}
\usepackage{xcolor}
\usepackage{colortbl} 
\usepackage{tikz}
\usepackage{mathtools}

\makeatletter
\def\adl@drawiv#1#2#3{%
        \hskip.5\tabcolsep
        \xleaders#3{#2.5\@tempdimb #1{1}#2.5\@tempdimb}%
                #2\z@ plus1fil minus1fil\relax
        \hskip.5\tabcolsep}
\newcommand{\cdashlinelr}[1]{%
  \noalign{\vskip\aboverulesep
           \global\let\@dashdrawstore\adl@draw
           \global\let\adl@draw\adl@drawiv}
  \cdashline{#1}
  \noalign{\global\let\adl@draw\@dashdrawstore
           \vskip\belowrulesep}}
\makeatother

\newcommand{\newtext}[1]{#1}

\newcommand{\cmmnt}[1]{\ignorespaces}

\newcommand{\symbolsecref}[1]{($\S$~\ref{#1})}

\definecolor{average}{HTML}{003300}
\newcommand{\avg}[1]{\textcolor{average}{#1}}

\newcommand{\semsup}[0]{\textsc{SemSup}}
\newcommand{\semout}[0]{\mathcal{O}^{\textrm{SS}}}
\newcommand{\supervised}[0]{\textsc{Sup}}
\newcommand{\ytrain}[1]{\mathcal{Y}_{\textrm{train}}}
\newcommand{\ytest}[1]{\mathcal{Y}_{\textrm{test}}}

\definecolor{abstractblue}{HTML}{0070C0}

\newcommand{\devise}[0]{\textsc{DeViSE}}
\newcommand{\gile}[0]{\textsc{GILE}}

\newcommand{\sota}[0]{\textsc{clip or t5}}
\newcommand{\sotanames}[0]{\textsc{clip or t5}}
\newcommand{\semsupbi}[0]{\textsc{SemSup (bi)}}
\newcommand{\semsupcoil}[0]{\textsc{SemSup (hybrid)}}
\newcommand{\biencodernames}[0]{\textsc{deep bi-enc}}

\newcommand{\lrap}[0]{\textsc{Lrap}}

\definecolor{mygreen}{RGB}{0,160,0}

\definecolor{myred}{RGB}{178,34,34}

\newcommand{\myparagraph}[1]{\paragraph{#1}}

\icmltitlerunning{\semsup{}: Semantic Supervision for Simple and Scalable Zero-shot Generalization}

\begin{document}

\twocolumn[
\icmltitle{\semsup{}: Semantic Supervision for Simple and Scalable \\ Zero-shot Generalization}



\icmlsetsymbol{equal}{*}




\begin{icmlauthorlist}
\icmlauthor{Austin W. Hanjie}{equal,yyy}
\icmlauthor{Ameet Deshpande}{equal,yyy}
\icmlauthor{Karthik Narasimhan}{yyy}
\end{icmlauthorlist}

\icmlaffiliation{yyy}{Department of Computer Science, Princeton University}

\icmlcorrespondingauthor{Austin W. Hanjie}{hjwang@cs.princeton.edu}
\icmlcorrespondingauthor{Ameet Deshpande}{asd@cs.princeton.edu}


\icmlkeywords{Machine Learning, ICML}

\vskip 0.3in
]



\printAffiliationsAndNotice{\icmlEqualContribution} 

\begin{abstract}
Zero-shot learning is the problem of predicting instances over classes not seen during training. One approach to zero-shot learning is providing auxiliary class information to the model.
Prior work along this vein have largely used expensive per-instance annotation or singular class-level descriptions, but per-instance descriptions are hard to scale and single class descriptions may not be rich enough.
Furthermore, these works have used natural-language descriptions exclusively, simple bi-encoders models, and modality or task-specific methods.
These approaches have several limitations: text supervision may not always be available or optimal and bi-encoders may only learn coarse relations between inputs and class descriptions.
In this work, we present \semsup{}, a novel approach that uses (1) a scalable multiple description sampling method which improves performance over single descriptions, (2) alternative description formats such as JSON that are easy to generate and outperform text on certain settings, and (3) hybrid lexical-semantic similarity to leverage fine-grained information in class descriptions. We demonstrate the effectiveness of \semsup{} across four datasets, two modalities, and three generalization settings. For example, across text and image datasets, \semsup{} increases unseen class generalization accuracy by 15 points on average compared to the closest baseline.\footnote{Code is available at: \url{https://github.com/princeton-nlp/semsup}}
\end{abstract}
\section{Introduction}

\begin{figure*}[t!]
    \centering
    \includegraphics[width=0.8\linewidth]{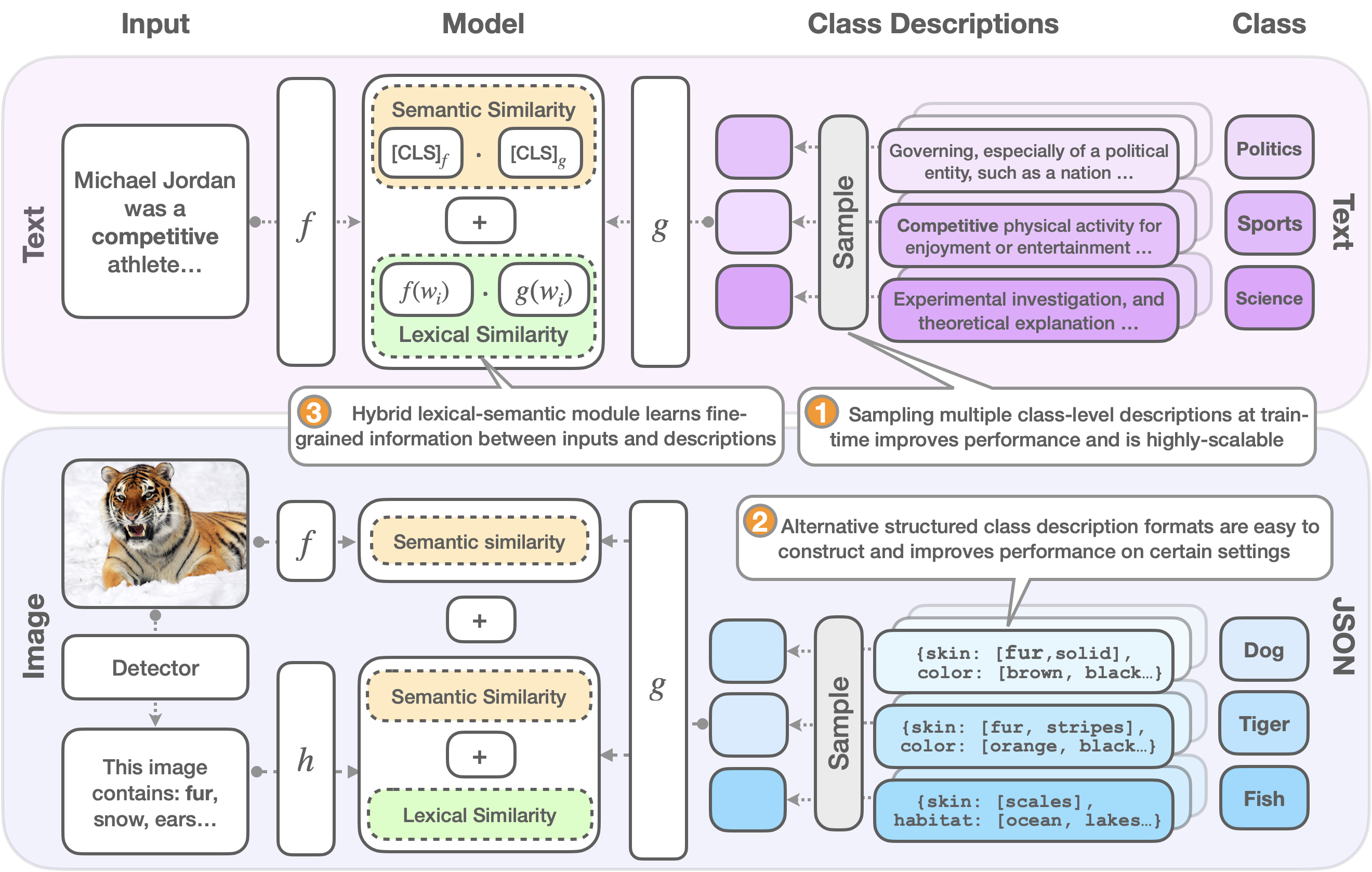}
    \caption{
    Our approach \semsup{} features three key innovations that improves zero-shot generalization: (1) multiple class-level descriptions that improve performance and are easy to scale, (2) alternate class description formats like structured JSON that are easy to generate automatically and can outperform text on certain settings, and (3) a hybrid lexical-semantic similarity model that learns fine-grained correspondence between inputs and class descriptions.
    Here $f$ is the input encoder, and $g$ is the class descriptions encoder.
    Note that the semantic and lexical similarity modules for the image case (bottom) are identical to the text case (top) and details are omitted for clarity.
    We run the image through an object detector and convert them to a text format for the lexical similarity module. 
    This figure shows two sample variants of \semsup{} models: (1) text inputs with text class descriptions (top) and (2) image input and structured JSON class descriptions (bottom).
    Any combination of input and output modalities can be used, and we evaluate all combinations in this work.
    }
    \label{fig:teaser}
\end{figure*}

Traditional supervised classification models are trained to map inputs from some feature space to a set of output classes which is fixed between training and inference. A longstanding goal of supervised learning is \textit{zero-shot generalization}, or the ability to classify instances over classes not seen during training \cite{chang2008importance, larochelle2008zero, lampert2009learning}. One way to achieve this generalization is integrating class semantics into the learning framework. Such semantic information can take the form of semantic codes~\cite{palatucci2009zero}, class name embeddings~\cite{socher2013zero,frome2013devise}, dictionary/wiki definitions~\cite{nam2016all,pappas2019gile}, and instance-level captions \cite{reed2016learning, radford2learning}.

While these works demonstrate the feasibility of class semantic information for the zero-shot paradigm, they often develop methods specific to modalities (e.g. image) or tasks (e.g. AWA2 \cite{xian2018goodbadugly} which has attribute annotations). Further, these works largely employ instance-level text annotations (e.g. captions) or single descriptions per class, use natural language (NL) text exclusively to convey class semantics, and model the input and class relations with simple bi-encoders~\cite{bai2009supervised}. These designs have multiple shortcomings: (1) single class-level descriptions are not semantically rich enough, and instance-level descriptions are hard to scale, (2) NL descriptions may not always be available or optimal for every setting, and (3) bi-encoders may only capture coarse correspondences between class descriptions and inputs~\cite{khattab2020colbert}.

In this work we introduce \textit{semantic supervision} (\semsup{}), a unified approach for zero-shot learning which achieves strong performance across multiple modalities and generalization settings (Figure~\ref{fig:teaser}). \semsup{} contains three crucial components: (1) A sampling strategy for usage of multiple class-level descriptions which significantly improves performance over single class descriptions, and requires up to 500$\times$ less annotated examples compared to instance-level supervision (on CIFAR). (2) The addition of alternative class description formats like structured JSON which are easy to generate automatically and provide superior performance on certain settings. (3) A hybrid semantic-lexical similarity module which leverages fine-grained lexical information \cite{gao2021coil} that improves generalization abilities.
Finally,
we introduce an automated class-description collection approach using search-engine scraping which can be used for any dataset and is highly scalable.

We evaluate \semsup{} across two text datasets (20NG, RCV1) and two image datasets (CIFAR100, AWA2), on both multiclass and multilabel classification, and across three generalization scenarios (unseen descriptions, unseen classes, and unseen superclasses). \semsup{} outperforms prior approaches by an average of 15 and 9 points on unseen class and superclass generalization respectively, while not giving up any performance compared to a traditional supervised baseline on seen classes (i.e. the standard supervised setup).
Further, it outperforms prompting based approaches on CLIP and T5 by over $10$ points on unseen classes, while being pretrained on significantly lesser data ($1/30$).

The capabilities of \semsup{} across tasks and modalities combined with the minimal data overhead afforded by our training scheme and automated description collection process suggests that \semsup{} can be a viable drop-in replacement for standard supervised learning. We also perform several ablation studies and analyses to identify the importance of various components in \semsup{}.

To summarize, we propose \semsup{}, an approach for zero-shot generalization which can be applied across modalities and tasks and improves generalization performance over prior approaches, while being simple and highly scalable. 

\section{Methodology}
\label{sec:methodology}

\newcommand{\model}[0]{$\mathcal{M}_{\theta}$}
\newcommand{\ppred}[0]{$P_{pred}(y|x_i)$}
\newcommand{\psup}[0]{$P_{\textrm{\supervised{}}}(y|x_i)$}
\newcommand{\psemsup}[0]{$P_{\textrm{\semsup{}}}(y|x_i)$}
\newcommand{\mcal}[1]{\mathcal{#1}}

\newcommand{\scenarioenum}[2]{\textbf{#1#2}}

\subsection{Background}
\label{sec:methodology:background}
Let $x_i$ and $\mcal{Y} = \{1, \dots, K\}$ denote the input and the categorical output. 
The model consists of (1) an input encoder $f(\cdot)$ and (2) an output matrix, whose $i^{\textrm{th}}$ row ($\mathcal{O}[i]$) is a representation for the $i^{\textrm{th}}$ class. Logits are obtained as $\mathcal{O}f(x_i)$. In standard supervised learning (\supervised{}), $\mathcal{O}$ is randomly initialized, and devoid of any semantic meaning. Prior zero-shot approaches inject semantics by representing classes using attributes~\cite{xian2018goodbadugly} class names~\cite{frome2013devise} or single descriptions~\cite{bujwid2021large,qiao2016less} using an output encoder $g(\cdot)$. However, single descriptions may not be semantically dense enough, NL may not always be available, and bi-encoders may only capture coarse information.


\subsection{Semantic Supervision}
\label{sec:methodology:semsup}
Our approach (\semsup{}) addresses the aforementioned shortcomings through:
(1) multiple semantically-rich class descriptions,
(2) flexible formats to describe classes (natural language and structured formats), and
(3) incorporating both lexical and semantic similarity. 

\paragraph{Multiple class descriptions} \semsup{} relies on multiple semantically rich descriptions to gain a thorough understanding of a class. These descriptions can reference different attributes to provide a holistic understanding of classes.
For example, the class \textit{sports} can be described in terms of its definition, examples, etymology and so on.

Formally, let $\mathcal{C} = \left ( \mathcal{C}_1, \dots, \mathcal{C}_K \right )$ be the descriptions for $K$ classes, where $\mathcal{C}_j = \{ c_{j}^1, \dots, c_{j}^L \}$ contain a set of $L$ descriptions describing the $j^{\textrm{th}}$ class.\footnote{Prior work is a special case where $\mcal{C}_j = \{c_j^1\}$} For each batch $b$ during training and testing, we dynamically construct the output matrix $\semout_b$ by sampling class descriptions $c_j$ uniformly and independently from $\mcal{C}_j$ for every class $j\in\mcal{Y}$:
\begin{align}
    \semout_b[j] = g(c_j)\qquad c_j \sim \mathrm{Unif}(\mcal{C}_j)\quad\forall j\in\mcal{Y}
\end{align}
At test-time, we predict the class corresponding to the class description with the highest softmax probability.\footnote{We found that changing random class description samples did not significantly change the overall accuracy on the evaluation sets.} Our approach can be flexibly applied to existing zero-shot methods and allows \semsup{} to leverage a large number of class descriptions with minimal data and compute overhead.

\paragraph{Alternative supervision formats} While text has been the de facto choice for class descriptions, for certain datasets describing classes in structured formats is easier than natural language.
For example, the class \textit{tiger} can be described as a JSON: \texttt{\{skin: [fur, stripes],color: [orange, black…\}}.
Compared to categorical class attribute vectors, these structured forms provide additional flexibility --- for instance, we can specify hierarchical information (color vs. orange), or free-form text formats (e.g. natural language definitions as a value). This information is difficult to incorporate using prior techniques with multihot vectors or word vector embeddings of attribute names (e.g. \citet{koh2020concept, demirel2017attributes2classname}).

\paragraph{Hybrid lexical-semantic similarity}
We propose two bi-encoder architectures
-- (1) \semsupbi{} which uses only semantic similarity and (2) \semsupcoil{} which uses a hybrid lexical-semantic similarity model.

For an input $x_i$, \semsupbi{} uses the product between the class description and input representation $\semout_bf(x_i)$ as class logits.\footnote{We add a projection matrix $\mathcal{P}$ to $f$ to ensure dimensions match between $f(\cdot)$ and $\semout_b$} 
While \semsupbi{} can learn semantic similarity, it ignores lexical cues which are strong indicators of compatibility.
For example, for the input \textit{``Michael Jordan was \underline{competitive}''} and class \textit{``sports''} with the description \textit{``A \underline{competitive} physical activity or game''}, the common word \textit{competitive} is important to identify similarity.
To incorporate this information, we propose \semsupcoil{} motivated by~\citet{gao2021coil}. \semsupcoil{} incorporates lexical similarity using the dot-product of the contextualized representations of lexically identical words between input and outputs.
Concretely, for input $x_i$ and class description $c_{j}$\footnote{Identical to the one used for $\semout_b$.}, let $f_{x_i}(w)$ and $g_{c_{j}}(w)$ be the contextualized representation of a common word $w \in x_i \cup c_{j}$ (e.g., \textit{competitive}). The logits for \semsupcoil{} are:
\begin{align}
    \semout_bf(x_i) + \sum_{w \in x_i\:\cup\:c_{j}} f_{x_i}(w_i)^\top g_{c_{j}}(w_i)
\end{align}
To utilize \semsupcoil{} for image inputs, we construct a natural language sentence by obtaining image annotations like the color, shape, or background of the object ($\bar{x}_i$).
Formally, let $h(\cdot)$ be the image annotation encoder,\footnote{Again with projection matrix $\mathcal{P}_I$ to ensure dimensions match} the \semsupcoil{} logits are given by the sum of the (1) image and description semantic similarity, (2) image annotation and description semantic similarity, and (3) image attributes and description lexical similarity:
\begin{align}
    \begin{split}
    \semout_bf(x_i) + g(c_{j})^\top \:h(\bar{x}_i) + \\
    \sum_{w \in \bar{x}_i\:\cup\:c_{j}} h_{\bar{x}_i}(w_i)^\top g_{c_{j}}(w_i)
    \end{split}
\end{align}
During training, we learn the input encoders $f(\cdot)$, $h(\cdot)$ and class description encoder $g(\cdot)$ together.

\paragraph{Usefulness of \semsup{}}
Intuitively, \semsup{} makes predictions over output choices by `reading' their `description' and deciding the best match for the input instance.
This is akin to training models to answer \textit{informative} multiple-choice questions, which provides several flexible use cases during inference:
(1) new unseen choices can be specified using corresponding descriptions,
(2) the number of choices can be dynamic,
and
(3) classes can be described in different ways based on end user preference,


\subsection{Collecting output supervision}
\label{sec:output_sup}

\semsup{} is not limited by output supervision format (e.g. class name or attribute vector).
We consider two kinds of descriptions -- (a) natural language and (b) structured (in JSON format), which we collect in a semi-automated fashion.
Our pipeline enables collecting descriptions of both modalities for any dataset with minimal effort. We present example descriptions in Table \ref{tab:examples}.
\begin{table}
\centering
\small
\resizebox{\linewidth}{!}{
\begin{tabular}{p{0.95\linewidth}}
\toprule
\rowcolor{gray!10} \multicolumn{1}{c}{\textbf{RCV1 (Consumer Prices)}} \\
\midrule
\underline{\textbf{Text:}} A consumer price index is a price index, the price of ...
\\
\cdashlinelr{1-1}
\underline{\textbf{Structured:}} \textit{\{\textbf{def.}: The consumer price index uses a basket of products ...
\textbf{related terms}: [`consumer price', `inflation', ...]
\}}\\
\midrule \rowcolor{gray!10} \multicolumn{1}{c}{\textbf{AWA2 (Killer whale)}} \\ \midrule
\underline{\textbf{Text:}} Orcas (Killer whales) are one of 35 species ...
\\ \cdashlinelr{1-1}
\underline{\textbf{Structured:}} \textit{\{\textbf{appendages}: [flippers, tail], \textbf{color}: [black, white] \textbf{habitat}: [arctic, coastal, ocean, ...
\}} \\ 
\bottomrule
\end{tabular}
}
\caption{
Example descriptions for a single class from one text (RCV1) and one image (AWA2) dataset.
}
\label{tab:examples}
\end{table}

\label{sec:setup:supervision}
\paragraph{Natural language supervision}
We collect class descriptions using queries: ``what is a \textit{class}'' and issuing them to two search engines \url{google.com} and \url{duckduckgo.com}.
We scrape the resulting preview snippets, automatically remove scraper artifacts and partial descriptions, and manually filter any remaining off-topic descriptions.\footnote{Omitting manual filtering does not significantly affect performance, see Appendix \ref{app:description_quality}}

\myparagraph{Structured supervision (JSON)}
For some datasets, supervision is easier to specify in a structured format.
We collect and provide supervision in a JSON format to evaluate its usefulness.
The JSONs contain attribute-value pairs for each class, with different datasets having their own attributes.
When class attributes are available (AWA2), we directly use them in the JSON (e.g., {\tt \{color:[orange, black]\}} for a tiger).
When class attributes are not available (RCV1, 20NG, CIFAR) we automatically construct the JSON by scraping dictionary definitions, hyponyms, and hypernyms from WordNet \cite{miller1995wordnet} and related terms from thesauri~\cite{PowerThesaurus}.
To increase description diversity, we automatically augment the descriptions by permutations and randomly dropping out values.

\paragraph{Generating image attributes}
To enable \semsupcoil{}'s use of lexical similarity on image datasets, we obtain image attributes from an object detector \cite{kolesnikov2020bigtransfer}, convert them to comma separated text, and append them to the template \textit{``This photo contains:''}.
We ensure that the list of attributes do not contain information about the ground truth class of the image by omitting detections that overlap with the class name, synonyms, or hyponyms.
Synonyms and hyponyms are removed with the help of WordNet synsets and hierarchies respectively.

Our use of pretrained image-detection systems follows recent works which chain multiple pretrained systems (including object detectors) to enable new capabilities not possible with individual components \cite{zeng2022socratic}.
\section{Experimental Setup}
\label{sec:experimental_setup}

\subsection{Evaluating \semsup{}'s generalization}
\begin{table}
\centering
\resizebox{\linewidth}{!}{
\begin{tabular}{p{0.15\linewidth}p{0.05\linewidth}p{0.25\linewidth}|p{0.05\linewidth}p{0.5\linewidth}}
\toprule
     & \multicolumn{2}{c}{\textbf{Classes}} & \multicolumn{2}{c}{\textbf{Class Descriptions}} \\ \midrule
    \rowcolor{gray!10} \multicolumn{5}{c}{\textbf{Train}} \\ \midrule
\textbf{S[0-3]} & $\mathcal{Y}$ & {\tt sharks, flatfish, frog} & \textit{} $\mathcal{C}$ & \textit{Sharks are a elasmobranch fish with a cartilaginous skeleton...} \\
\midrule \rowcolor{gray!10} \multicolumn{5}{c}{\textbf{Test}} \\ \midrule
\textbf{S0} & $\mathcal{Y}$ & {\tt sharks, flatfish, frog} & $\mathcal{C}$ & \textit{Sharks are a elasmobranch fish with a cartilaginous skeleton...}\\\cmidrule{1-5}
\textbf{S1} & $\mathcal{Y}$ & {\tt sharks, flatfish, frog} & $\mathcal{C}'$ & \textit{A predatory fish, modern sharks are also known as Selachimorpha...}\\\cmidrule{1-5}
\textbf{S2} & $\mathcal{Y}'$ & {\tt lions, ants}& $\mathcal{C}_{\mathcal{Y}'}$ & \textit{mammal that lives in the prairies...} \\\cmidrule{1-5}
\textbf{S3} & $\mathcal{Z}$ & {\tt fishes, reptiles} & $\mathcal{C}_{\mathcal{Z}}$ & \textit{fishes live in water and breath through gills...}\\
\bottomrule
\end{tabular}
}
\caption{
Illustration for classes and descriptions on scenarios 0-3 (S[0-3]).
Scenario 0 is the standard classification setup where classes remain fixed between train and test. In Scenario 1, classes are fixed but the class descriptions change. In Scenarios 2 and 3, we generalize to unseen classes and superclasses respectively.
}
\label{tab:scenarios}
\end{table}

We test~\semsup{}'s generalization under four scenarios at inference time (Table \ref{tab:scenarios}):
\\
\textbf{(S0)} Performance over unseen \textit{instances} of seen classes with seen  descriptions. This is equivalent to the standard supervised learning setup with in-distribution examples.
\\
\textbf{(S1)} Generalization to unseen instances and unseen \textit{descriptions} of classes it has previously seen during training.
\\
\textbf{(S2)} Generalization to unseen classes (specified using unseen descriptions) -- which is the common zero-shot learning setup~\cite{larochelle2008zero}.
\\
\textbf{(S3)} Generalization to unseen instances of superclasses, which are at a higher level of conceptual granularity than training classes.
This setting evaluates models' understanding of class hierarchies and relationships.

\subsection{Datasets}
\label{sec:setup:datasets}


We evaluate~\semsup{} on four diverse datasets  -- three of them have a hierarchical organization of classes and one which has annotated attributes (AWA2).
Except for RCV1 below, the datasets are multi-class classification datasets. Detailed dataset information are presented in Appendix~\ref{sec:dataset}.
\\
\textbf{1. 20 Newsgroups} (20NG)~\cite{lang1995newsgroups} consists of newsgroup documents of correspondences between online users. 
\\
\textbf{2. CIFAR-100}~\cite{krizhevsky2009cifar} contains images ranging from animals to household objects from Tiny Images \cite{torralba2008tinyimages}.
\\
\textbf{3. Animals with Attributes 2} (AWA2) \cite{xian2018goodbadugly} is an animal classification dataset which includes 85 binary animal attributes (e.g. \textit{fur}, \textit{swims}).
\\
\textbf{4. RCV1}~\cite{lewis2004rcv1} is a \textit{multi-label} news classification dataset over niche news classes (e.g., \textit{bond markets} and \textit{credit ratings}). We use the provided class hierarchy to construct superclasses and evaluate using label ranking average precision (\lrap{})~\cite{pappas2019gile}.

\subsection{Models}
\label{sec:setup:models}
For text datasets, we encode input features ($f(\cdot)$) using BERT-small~\cite{turc2019}. For image datasets, we use the patch32 Vision Transformer \cite{dosovitskiy2020vit} pretrained on ImageNet \cite{deng2017image}. Unless specified, we use the {\tt [CLS]} representation of both models. We train end-to-end with cross-entropy and binary cross-entropy loss for multiclass multi-label datasets respectively.

\myparagraph{Baselines}
We evaluate five baselines from prior work that differ in the class description representations ($g(\cdot)$):
\\
(1) \textsc{Sup} is a standard supervised baseline which uses an output matrix for classification and hence ignores any semantic class information.
\\
(2) \devise{}~\cite{frome2013devise} uses GLoVe~\cite{pennington2014glove} vectors of class names to represent the classes.
\\
(3) \gile{}~\cite{pappas2019gile} uses GLoVe mean bag-of-word-vectors of class descriptions and adds a $\mathrm{tanh}$ activation. While \citet{pappas2019gile} use a single class description for each class, we train \gile{} with multiple descriptions for fair comparison.
\\
(4) \biencodernames{} uses BERT-small to encode class names using a template (the class is / a photo of a {\tt [CLSNAME]}).
\\
(5) {\sc CLIP and T5}: We prompt CLIP~\cite{radford2learning} and T5-large~\cite{raffel2020exploring} using (1) class names and (2) text descriptions.
T5 is prompted in a cross-encoder setup~\cite{humeau2019poly,xue2021mt5}.
These models are significantly larger and pre-trained on orders of magnitude more data when compared to \semsup{} (30$\times$, 50$\times$).
Furthermore, CLIP is likely trained on the test set of the datasets we consider~\cite{radford2learning}, but we include it for the sake of completeness.
We still outperform these models on most settings and datasets.

\paragraph{\semsup{}}
For our \textbf{\semsup{}} models, we encode output features ($g(\cdot)$) using representations from a pretrained BERT-small model~\cite{turc2019} for both text and JSON descriptions.
We experiment with both \semsupbi{} and \semsupcoil{} described in~\symbolsecref{sec:methodology:semsup}.

\section{Results}
\label{sec:results}

\begin{table}[t]
    \centering
    \small
    \resizebox{\linewidth}{!}{%
    \begin{tabular}{lccccc@{}}
    \toprule
     \textbf{Model} & \textbf{RCV1} & \textbf{20NG} & \textbf{CIFAR} & \textbf{AWA} & \textbf{Avg}\\\midrule
     \textsc{Sup} & 95.5 & 92.2 & 89.2 & 95.9 & 93.2 \\
    \midrule \rowcolor{gray!10} \multicolumn{6}{c}{\textbf{Natural language descriptions}} \\ \midrule
    \rowcolor{gray!10} \multicolumn{6}{l}{\textbf{Class names}} \\
    \devise{} & 90.7 & 91.7 & 87.3 & 95.7 & 91.3\\
    \biencodernames{}\qquad\qquad\qquad & \textbf{95.7} & 91.7 & 88.0 & \textbf{95.8} & 92.8\\
    \rowcolor{gray!10} \multicolumn{6}{l}{\textbf{Descriptions}} \\
    \gile{} & 90.6 & 92.1 & 87.3 & 95.6 & 91.4\\
    \semsupbi{} & 95.5 & 92.2 & \textbf{88.6} & 96.1 & \textbf{93.1}\\
    \semsupcoil{} & \textbf{95.6} & \textbf{92.5} & 87.7 & \textbf{95.9} & 92.9\\ \midrule
    \rowcolor{gray!10} \multicolumn{6}{c}{\textbf{JSON descriptions}} \\ \midrule
    \gile{} & 94.8 & 91.6 & 87.4 & 95.4 & 92.3\\ 
    \semsupbi{} & \textbf{95.7} & 92.0 & \textbf{88.2} & \textbf{96.1} & \textbf{93.0}\\
    \semsupcoil{} & \textbf{95.6} & \textbf{92.2} & 87.1 & 95.9 & 92.7\\
    \bottomrule
    \end{tabular}%
    }
    \caption{
        Accuracy of models on unseen instances using descriptions identical to those used during training (scenario 0). \textsc{Sup} is a supervised baseline using the same input encoder with a learned output matrix. On all datasets, \semsup{} can match the performance of \textsc{sup}, with an accuracy within 0.2 points averaged over datasets.
        Numbers are averaged over three seeds. 
    }
    \label{tab:scen_0}
\end{table}
\newcommand{\unseentable}[0]{\textbf{\textsc{UN}}}
\newcommand{\seentable}[0]{\textbf{\textsc{S}}}

\begin{table*}[t]
    \centering
    \small
    \resizebox{\linewidth}{!}{%
    \begin{tabular}{lccccc|ccccc|ccccc@{}}
    \toprule
    \multicolumn{1}{l}{\multirow{3}{*}{\textbf{Model}}} & \multicolumn{5}{c}{\textbf{Unseen descriptions (S1)}} & \multicolumn{5}{c}{\textbf{Unseen classes (S2)}} & \multicolumn{4}{c}{\textbf{Unseen superclasses (S3)}} \\
    \cmidrule(lr){2-6} \cmidrule(lr){7-11} \cmidrule(lr){12-15}
    \multicolumn{1}{c}{} & {\textbf{RCV1}} & {\textbf{20NG}} & {\textbf{CIFAR}} & {\textbf{AWA}} & \textbf{\avg{Avg}} &\textbf{RCV1} & \textbf{20NG} & \textbf{CIFAR} & \textbf{AWA} & \textbf{\avg{Avg}} & \textbf{RCV1} & \textbf{20NG} & \textbf{CIFAR} & \textbf{\avg{Avg}} \\
    \midrule
    \rowcolor{gray!10} \multicolumn{15}{c}{\textbf{Natural language descriptions}} \\ \midrule
    \rowcolor{gray!10} \multicolumn{15}{l}{\textbf{Class names}} \\
    \sotanames{} & 5.5 & 37.5 & 41.1 & 90.0 & \avg{43.5} & 15.8 & 29.9 & \textbf{82.3} & \textbf{97.3} & \avg{56.3} & 31.4 & 27.5 & 54.2 & \avg{37.7} \\
    \devise{} & 51.7 & 86.0 & 68.2 & 71.1 & \avg{69.3} & 27.1 & 70.2 & 61.4 & 17.3 & \avg{44.0} & 47.0 & 78.9 & 53.0 & \avg{59.6}\\
    \biencodernames{} & 45.1 & 73.2 & 67.9 & 65.3 & \avg{62.9} & 44.6 & 72.4 & 68.9 & 16.5 & \avg{50.6} & 56.1 & 77.8 & 59.5 & \avg{64.5}\\
    \rowcolor{gray!10} \multicolumn{15}{l}{\textbf{Descriptions}} \\
    \sota{} & 5.5 & 14.0 & 34.4 & 84.6 & \avg{34.6} & 11.7 & 17.5 & 78.2 & 95.3 & \avg{50.7} & 21.9 & 27.0 & 53.6 & \avg{34.2} \\
    \gile{} & 53.0 & 88.5 & 67.6 & 70.7 & \avg{70.0} & 35.2 & 69.2 & 72.1 & 28.0 & \avg{51.1} & 46.2 & 81.3 & 64.1 & \avg{63.9}\\
    \semsupbi{} & \textbf{90.8} & 91.6 & \textbf{87.2} & \textbf{95.8} & \textbf{\avg{91.4}} & 48.0 & 70.4 & 74.6 & 35.0 & \avg{57.0} & 56.2 & \textbf{85.5} & 68.1 & \avg{69.9}\\
    \semsupcoil{} & 89.0 & \textbf{92.1} & 85.8 & 95.4 & \avg{90.6} & \textbf{59.2} & \textbf{89.3} & 80.8 & 39.8 & \textbf{\avg{67.3}} & \textbf{66.3} & 81.6 & \textbf{69.1} & \textbf{\avg{72.3}}\\ \midrule
    \rowcolor{gray!10} \multicolumn{15}{c}{\textbf{Structured JSON descriptions}} \\ \midrule
    \gile{} & 57.5 & 91.5 & 87.2 & 95.4 & \avg{82.9} & 34.1 & 69.8 & 75.2 & 33.8 & \avg{53.2} & 48.9 & \textbf{81.0} & 71.9 & \avg{67.3}\\ 
    \semsupbi{} & \textbf{84.6} & 91.9 & \textbf{88.0} & \textbf{96.0} & \textbf{\avg{90.1}} & 46.1 & 70.4 & 73.3 & \textbf{46.1} & \avg{59.0} & 54.0 & 80.0 & \textbf{72.8} & \textbf{\avg{68.9}}\\
    \semsupcoil{} & 83.5 & \textbf{92.2} & 86.9 & 95.8 & \avg{89.6} & \textbf{62.8} & \textbf{85.6} & \textbf{82.2} & 42.5 & \textbf{\avg{68.3}} & \textbf{57.3} & 75.3 & 72.0 & \avg{68.2}\\
    \bottomrule
    \end{tabular}%
    }
    \caption{
        Accuracy of models on unseen descriptions (S1), unseen classes (S2), and unseen superclasses (S3) using natural language (NL) class names and text, and structured JSON descriptions. \semsupcoil{} consistently outperforms other approaches on NL, beating the next best model by up to 10 points on S2 and 2 points on S3, which emphasizes the importance of its hybrid lexical-semantic similarity.
        \semsupcoil{} also outperforms \sota{} on average, even though the latter are pre-trained on significantly more data.
        For JSON descriptions, \semsup{} models outperform \gile{}, and \semsupcoil{} significantly outperforms \semsupbi{} on S2 (10 points) while being close in S1 and S3.
        CLIP is used for image datasets and T5 for text datasets.
        Numbers are averaged over three seeds.
    }
    \label{tab:all_results}
\end{table*}

\paragraph{\semsup{} matches \textsc{Sup} on in-distribution examples} We present the accuracy of models on unseen instances, but seen descriptions in Table \ref{tab:scen_0}.
On average across datasets, \semsupcoil{} and \semsupbi{} are within 0.3 percentage points of a standard supervised baseline (\textsc{Sup}). This is in contrast to \devise{} and \gile{}, which are 2 points worse than the supervised baseline, likely due to their fixed description encoders (GLoVE).
This shows that \semsup{} models are a viable drop-in replacement for standard supervision, and performance on the traditional classification settings on in-distribution examples is not hurt when compared to the supervised baseline.

\paragraph{\semsup{} exhibits strong generalization to scenarios with unseen elements} The performance of models on scenarios 1-3, is presented in Table \ref{tab:all_results}. Across all datasets, \semsupcoil{} and \semsupbi{} achieve the highest average performance on unseen descriptions (S1), beating the next best model (\gile{}) by up to 20 percentage points. Similar trends hold for unseen classes (S2) and unseen superclasses (S3), with \semsup{} and \semsupbi{} beating \gile{} and a bi-encoder class names baseline (\biencodernames{}) by up to 15 points on S2, and 8 points on S3 averaged across datasets. Across generalization scenarios and both image and text modalities, our \semsup{} models significantly outperform baselines, demonstrating that our approach is general-purpose and effective across a wide range settings.

\paragraph{Hybrid lexical-semantic similarity improves generalization} On scenarios 1-3, our \semsupcoil{} variant with lexical-semantic similarity consistently outperforms \semsupbi{}. For example, averaged across datasets, \semsupcoil{} outperforms \semsupbi{} by 10 percentage points on unseen classes and 2.4 points on unseen superclasses. These results suggest that lexical information can help models better leverage the semantic information in the class descriptions. In particular, for image datasets, lexical matching on image annotations can provide grounded information that would otherwise have to be learned from scratch. For example, the annotation ``farm'' may be helpful for identifying livestock classes. In section \ref{sec:results:ablation}, we conduct ablations to confirm that the strong performance of \semsup{} is not solely a product of strong image annotations. 

\newcommand{\reffortable}[0]{{\scriptsize $\leftarrow\textrm{\texttt{Reference}}$}}
\newcommand{\multirowtwo}[1]{\multirow{2}{*}{#1}}
\newcommand{\largestar}[0]{$\mathlarger{\mathlarger{\mathlarger{\star}}}$}

\begin{table*}[t]
    \centering
    \small
    \resizebox{\linewidth}{!}{%
    \begin{tabular}{lccccc|ccccc|ccccc@{}}
    \toprule
    \multicolumn{1}{l}{\multirow{3}{*}{\textbf{Descriptions}}} & \multicolumn{5}{c}{\textbf{Unseen descriptions (S1)}} & \multicolumn{5}{c}{\textbf{Unseen classes (S2)}} & \multicolumn{4}{c}{\textbf{Unseen superclasses (S3)}} \\
    \cmidrule(lr){2-6} \cmidrule(lr){7-11} \cmidrule(lr){12-15}
    \multicolumn{1}{c}{} & {\textbf{RCV1}} & {\textbf{20NG}} & {\textbf{CIFAR}} & {\textbf{AWA}} & \textbf{Avg} &\textbf{RCV1} & \textbf{20NG} & \textbf{CIFAR} & \textbf{AWA} & \textbf{Avg} & \textbf{RCV1} & \textbf{20NG} & \textbf{CIFAR} & \textbf{Avg} \\
    \midrule\rowcolor{gray!10} \multicolumn{15}{c}{\textbf{\semsupbi{}}} \\ \midrule
    Concat-10 & 73.1 & 83.1 & 79.9 & 61.5 & 74.4 & 39.1 & 57.2 & 67.3 & 27.3 & 47.7 & 50.1 & 68.8 & 60.7 & 59.9 \\
    $n=1$ & 38.3 & 70.7 & 58.6 & 28.6 & 49.1 & 35.6 & 62.8 & 61.4 & 23.1 & 45.7 & 46.4 & 66.7 & 61.8 & 58.3 \\
    $n=5$ & 81.9 & 91.5 & 86.8 & 95.5 & 88.9 & \textbf{49.3} & 66.3 & 72.1 & \textbf{35.7} & 55.9 & \textbf{61.7} & 83.5 & \textbf{72.0} & \textbf{72.4 }\\
    $n=10$ & \textbf{{90.8}} & \textbf{91.6} & \textbf{{87.2}} & \textbf{{95.8}} & \textbf{91.4} & 48.0 & \textbf{70.4} & \textbf{74.6} & 35.0 & \textbf{57.0} & 56.2 & \textbf{85.5} & 68.1 & 69.9\\
    \midrule\rowcolor{gray!10} \multicolumn{15}{c}{\textbf{\semsupcoil{}}} \\ \midrule
    Concat-10 & 78.3 & 81.7 & 68.8 & 59.5 & 72.1 & 50.1 & 84.4 & 74.3 & 31.9 & 60.0 & 57.6 & 67.0 & 58.0 & 60.9 \\    
    $n=1$ & 53.9 & 85.1 & 59.4 & 38.2 & 59.2 & 50.3 & 89.5 & 73.5 & \textbf{43.5} & 64.2 & 58.9 & 75.5 & 61.1 & 65.2 \\
    $n=5$ & 80.2 & 90.9 & 85.1 & 95.1 & 87.8 & \textbf{59.7} & \textbf{90.4} & 79.6 & 41.5 & \textbf{67.8} & \textbf{66.3} & 79.9 & \textbf{70.5} & 72.2 \\
    $n=10$ & \textbf{89.0} &\textbf{ 92.1} & \textbf{85.8} & \textbf{95.4} &\textbf{ 90.6} & {59.2} & 89.3 & \textbf{80.8} & 39.8 & 67.3 & \textbf{{66.3}} & \textbf{81.6} & {69.1} & \textbf{72.3}\\
    \bottomrule
    \end{tabular}%
    }
    \caption{
        Performance of \semsupcoil{} and \semsupbi{} when using different number of natural language training descriptions. $n=k$ indicates usage of $k$ descriptions per class during training, and Concat-$k$ concatenates the same $k$ descriptions and uses it as the sole description. On nearly all settings we see an increase in accuracy as we increase the number of sampled descriptions, indicating that multiple class-level descriptions is a scalable way to increase performance. Numbers averaged over 3 seeds.
    }
    \label{tab:num_descr}
\end{table*}

\paragraph{Supervision format can play an important role in generalization performance} From Table \ref{tab:all_results}, we find that using JSON instead of natural language improves unseen class accuracy by up to 11 percentage points on image classification (AWA) and 3 percentage points on text classification (RCV1). Averaged across datasets, JSON descriptions improves the unseen class accuracy of models between 1-2 percentage points. These results suggest that structured descriptions are a viable alternative to natural language, and may be more advantageous in certain settings. Compared to natural language, JSON could be more information dense for a given context length, and may be easier to automatically construct given metadata information. Both our \semsup{} models are able to flexibly use both description formats as necessary. Similar to works on prompt engineering, our results also suggest that task performance can be boosted via alternate description formats, and we encourage future works to explore this direction.
 
\paragraph{\semsup{} outperforms CLIP/T5 models} We observe from Table \ref{tab:all_results} that T5 consistently underperforms all other models. This gap in performance is likely due to class descriptions rarely following the corresponding text sequences in the pretraining data. In contrast, CLIP fares much better, achieving the strongest results on unseen classes on AWA2. However, CLIP is trained on orders of magnitude more data
and it is unclear if the unseen classes here which feature common animals are truly unseen by CLIP. Furthermore CLIP underperforms \semsupcoil{} by 15 points on unseen superclasses and only achieves a 41\% classification accuracy on CIFAR100 despite CIFAR100 appearing in the CLIP pretraining data \cite{radford2learning}. This may be due to the higher level descriptions of superclasses and the low resolution of CIFAR images which may be out of distribution with the rest of the CLIP pretraining. Despite the impressive capabilities of CLIP, these shortcomings highlight the need for methods trained for task-specific purposes.

\section{Analysis}
\label{sec:results:ablation}
\newcommand{\layers}[1]{$\mathbf{\left (L=#1\right )}$}

In this section we study the effect of multiple descriptions and image annotations on performance and conduct qualitative evaluations of input and output encoder embeddings.

\paragraph{Effect of Number of descriptions}
\label{sec:no_of_desc}
In Table \ref{tab:num_descr} we evaluate the performance of \semsupcoil{} and \semsupbi{} trained with $n=\{1,5,10\}$ descriptions. The $n=1$ setting is similar to prior works. Our sampling method using $n=10$ descriptions consistently outperforms $n=1$ by as many as 50 points on unseen class descriptions, 13 points on unseen classes, and 20 points on unseen superclasses.

To determine whether the improvements from $n=1$ to $n=10$ is due to our sampling scheme or increases in information content, we concatenate all of the $n=10$ descriptions for each class into one long description at training time (Concat-$10$). This ``super-description'' is similar to using entire Wikipedia paragraphs as class descriptions \cite{bujwid2021large}. The Concat-$10$ approach significantly underperforms sampling $n=10$ descriptions by between 7 and 18 percentage points on average, suggesting that sampling descriptions is important for learning robust discriminative features.

Our results indicate that sampling class descriptions is a scalable and effective alternative to instance-level supervision for increasing performance. For instance, following \citet{reed2016learning} who use annotators to collect 10 descriptions per training image, we would need 500,000 annotations on CIFAR100. Using our approach, we only need $\sim$1000 descriptions -- a 500$\times$ decrease.

\paragraph{Both images and their annotations are crucial for performance.}

\begin{table}[t]
\centering
\resizebox{\linewidth}{!}{
\begin{tabular}{lcccccc}
\toprule
\multirow{2}{*}{\textbf{Variant}} &\multicolumn{1}{l}{\multirow{2}{*}{\textbf{Modality}}} & \multicolumn{2}{c}{AWA2} & \multicolumn{3}{c}{CIFAR}\\
\cmidrule(lr){3-4} \cmidrule(lr){5-7}
& & \textbf{S1} & \textbf{S2} & \textbf{S1} & \textbf{S2} & \textbf{S3}\\ \midrule
{\small \semsupcoil{}} & JSON & \textbf{95.4} & \textbf{42.5} & \textbf{86.9} & 82.2 & \textbf{72.0}\\
{\small \semsup{} (Text-only)}  & JSON & 81.3 & 25.8 & 58.6 & \textbf{83.4} & 63.4\\
\hline
{\small \semsupcoil{}}& NL & \textbf{95.4} & \textbf{{39.8}} & \textbf{85.8} & \textbf{{80.8}} & \textbf{{69.1}}\\
{\small \semsup{} (Text-only)} & NL & 80.9 & 35.8 & 58.6 & 75.4 & 62.3\\
\bottomrule
\end{tabular}
}
\caption{  
    Accuracy of \semsupcoil{} and \semsup{} (Text-only) on AWA2 and CIFAR-100, averaged over 3 seeds. \semsup{} (Text-only) is an ablated version of \semsupcoil{} that uses text image annotations only without image input. \semsup{} (Text-only) significantly underperforms our model, indicating that both  \semsupcoil{}'s performance is not a result of both the image annotations and the image itself.
}
\label{tab:image_ablations}
\end{table} In Table \ref{tab:image_ablations}, we evaluate an ablated variant of \semsupcoil{} called \semsup{} (Text-only), that uses the textual image annotations, but not the image input. Performance of this text-only model is significantly worse than \semsupcoil{}; for example, on AWA2 (S2) accuracy drops by $17$ points. Conversely, using the images only without their annotations (i.e. \semsupbi{}) also results in significant drops in performance. For instance, removing annotations drops accuracy by 10 percentage points on unseen class generalization.
These results suggest that annotations do not directly reveal the identity of the underlying class; rather, the performance improvements of \semsupcoil{} over other models hinges on both (1) lexical-semantic similarity between textual image attributes and class descriptions, as well as (2) semantic similarity between raw images and class descriptions.
A qualitative analysis of the image annotations reveal that they often provide auxiliary information about the ground truth class that may be referenced in the class descriptions (e.g. {\tt prairie} for the class {\tt lion}).
Our results suggest that using a pretrained detection system in tandem with \semsup{} can provide additional grounded information that may increase model performance.


\myparagraph{Qualitative Analysis of Embedding Space}
\label{analysis:qualitative}
\begin{figure}[t]
    \centering
    \includegraphics[width=0.9\linewidth]{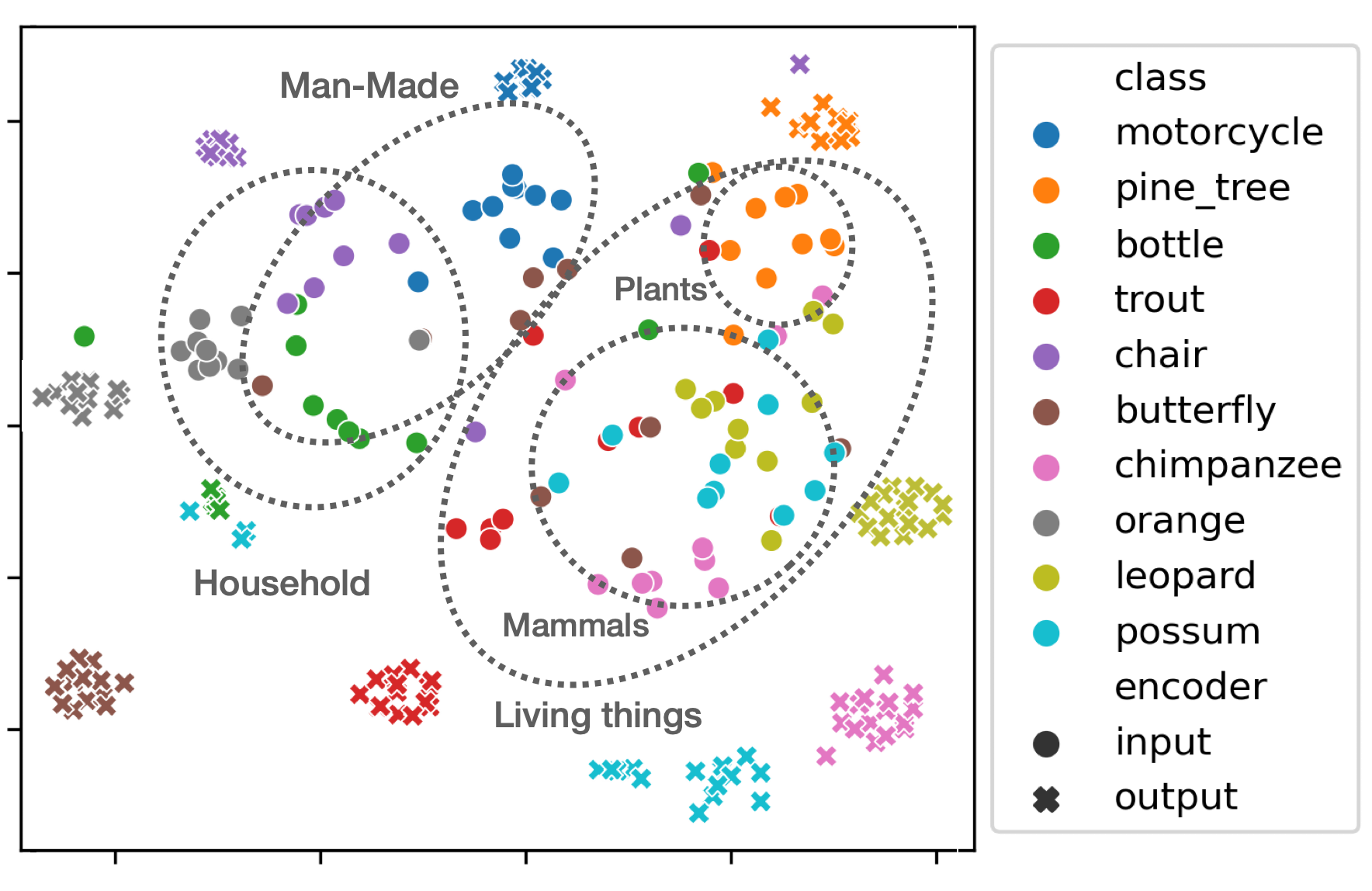}
    \caption{\newtext{T-SNE plot for input and class description encoder embeddings for held out classes on CIFAR-100 for \semsupbi{}.
    Each point is either an input instance embedding (dot) or a class description embedding (cross).
    Input and description embeddings belonging to the same class are closer together and classes belonging to similar concepts are grouped (e.g. man-made objects).}}
    \label{fig:tsne}
\end{figure}
To assess whether the learned input and output embeddings exhibit meaningful structures, we plot their t-SNE visualizations for \textit{unseen} classes on CIFAR-100 in Figure \ref{fig:tsne} after standardizing and normalizing the embeddings.
We observe that even for unseen classes, input and output embeddings corresponding to the same classes are closer together. Moreover, classes with similar conceptual overlap are grouped together. For example, living things clustered, with sub-clusters inside corresponding to mammals, plants and fishes.
This shows that \semsup{} gains semantic understanding of the classes, which may explain its strong performance.
\section{Related Work}
\label{sec:related_work}

\paragraph{Zero-shot learning using auxiliary information}
Zero-shot learning~\cite{larochelle2008zero} requires a model to classify instances into classes not seen during training, which necessitates representing classes using auxiliary information.
Several works encode class information using attributes  ~\cite{akata2015label,demirel2017attributes2classname,lampert2009learning,koh2020concept}, prototypes~\cite{snell2017prototypical,vinyals2016matching}, or word embeddings of class names/descriptions~\cite{socher2013recursive, frome2013devise, dauphin2014zero, reed2016learning, mittal2021decaf,wang2018joint,zhang2018multi, nam2016all,ba2015predicting,chang2008importance,pappas2019gile,bujwid2021large,qiao2016less}.

These prior works are developed and evaluated on single modalities or specific tasks, use simple bi-encoders or bag of vectors, and exclusively use natural language class descriptions.
In contrast, \semsup{} is task and modality agnostic and we evaluate on both image and text inputs, uses a hybrid lexical-semantic model for learning fine-grained correspondences, and uses alternate sources of semantic supervision aside from text, like structured descriptions.
This allows \semsup{} to outperform all baselines considered.

To the best of our knowledge, our work (\semsup{}) is the first to propose the use of multiple class descriptions to gain a fine-grained understanding of the class.
Some prior works have used instance-level supervision to learn fine-grained grounded features for zero-shot learning \cite{reed2016learning, radford2learning}.
For example, training CLIP~\cite{radford2learning} requires multiple image-specific captions, and as a result, the number of descriptions scales linearly with the number of instances in the dataset, thus requiring a large number of annotations.
In contrast, our method uses \textit{class}-level descriptions, and is significantly cheaper because the number of classes are typically orders smaller than the number of instances, and the same descriptions can be re-used for multiple instances of the same class.
\semsup{}'s data-collection pipeline provides an inexpensive way to collect high-quality descriptions automatically using search-engine scraping.
\citet{bujwid2021large} use single long descriptions which can be as big as a paragraph, but such lengthy supervision is not always available for all classes.
\semsup{} on uses multiple shorter descriptions to capture fine-grained semantics which outperforms the former.

\paragraph{Zero-shot prompting of pre-trained models}
A recent area of research has used prompting large pre-trained models for classification~\cite{liu2021pre,brown2020language,flanwei,sanh2022multitask,radfordlanguage,raffel2020exploring,schick2021s,gao2021making}, and models such as T5~\cite{raffel2020exploring}, GPT-3~\cite{brown2020language}) and CLIP~\cite{radford2learning} can be used to directly predict classes in a zero-shot fashion.
While large pre-trained models can achieve respectable zero-shot performance on a wide range of settings, they can completely fail on certain tasks outside the pre-training distribution (as we show in \symbolsecref{sec:results}), and \semsup{} outperform them while having significantly less parameters.
\semsup{} is orthogonal to prompting pre-trained models and provides a way to efficiently fine-tune them for tasks they fail on.

\paragraph{Learning with auxiliary language descriptions}
Related lines of work that inject semantic information into the learning framework include papers that incorporate auxiliary explanations~\cite{srivastava2017joint,srivastava2018zero,murty2020expbert,hancock2018training,mu2020shaping,liang2020alice,clarke2010driving,fidler2017teaching,mitchell1986explanation,dejong1986explanation} and labeling functions for weak supervision~\cite{hancock2018training,ratner2017snorkel,lison2020named,safranchik2020weakly,varma2018snuba,mayhew2019named}. Both these lines of work improve few-shot learning, whereas we focus on generalizing to new classes. Our work is also related to studies that learn classifiers~\cite{andreas2018learning}, reinforcement learning agents~\cite{branavan2010reading,branavan2012learning,denil2017programmable,andreas2018learning,zhong2019rtfm,narasimhan2018grounding,sharma2021skill,hanjie2021grounding}, and programs~\cite{acquaviva2021communicating,wong2021leveraging,desai2016program} by ``reading'' natural language descriptions of the task. While these works improve generalization or performance on specific domains and settings, it is unclear how to extend these approaches to general supervised learning setups.
\section{Conclusion}
\label{sec:conclusion}

We have presented \semsup{}, a unified approach for zero-shot learning that works across text and image modalities. \semsup{} provides a scalable drop-in replacement to any supervised classification problem, while solving key limitations of prior work using auxiliary information with three key features: (1) a scalable multiple description sampling method which improves performance over single descriptions (2) alternative JSON description formats that are easy to generate and outperform text on certain settings, and (3) lexical matching to leverage fine-grained information in class descriptions. \semsup{} achieves strong results across image and text modalities on a wide range of generalization settings, while remaining highly scalable in terms of both data and compute. We hope that our work will make zero-shot learning a more compelling alternative to supervised learning, and believe that future work can develop stronger models, explore a wider variety of semantic supervision, and apply this technique to other domains.

\bibliography{papers}
\bibliographystyle{icml2023}

\newpage
\appendix
\onecolumn
\section{Descriptions}
\label{sec:example_descriptions}

\begin{table*}[t]
\centering
\resizebox{\textwidth}{!}{%
\begin{tabular}{@{}llp{0.80\linewidth}}
\toprule
\textbf{Dataset} &
  \textbf{Class} &
  \textbf{Description} \\ \midrule
\multirow{3}{*}{\textbf{RCV1}} &
  \multirow{3}{*}{\textbf{Consumer prices}} &
  \underline{\textbf{Text:}} A consumer price index is a price index, the price of a weighted average market basket of consumer goods and \textbf{}services purchased by households. \\
  \cdashlinelr{3-3}
 &
   &
  \underline{\textbf{Structured:}} \textit{\{\textbf{definition}: The consumer price index uses a basket of products ranging from gasoline and health care to groceries and rents. \textbf{related terms}: [`consumer price', `consumer items', `hicp inflation', `indirect material', `consumer inflation rate', `consumer inflation']\}} \\ \midrule
\multirow{2}{*}{\textbf{20 NG}} &
  \multirow{2}{*}{\textbf{Cryptography}} &
  \underline{\textbf{Text:}} Cryptography is the study and practice of sending secure, encrypted messages between two or more parties \\ \cdashlinelr{3-3}
 &
   &
  \underline{\textbf{Structured:}} \textit{\{\textbf{examples}: [Modern cryptography is heavily based on mathematical theory…], \textbf{hypernyms}: [communication, security,…], \textbf{definition}: the science of analyzing and deciphering codes…, \textbf{hyponyms}: [encryption]\}} \\ \midrule
\multirow{2}{*}{\textbf{CIFAR}} &
  \multirow{2}{*}{\textbf{Flatfish}} &
  \underline{\textbf{Text:}} A category of fish that are characterized by their narrow bodies that are flat and oval-shaped. \\ \cdashlinelr{3-3}
 &
   &
  \underline{\textbf{Structured:}} \textit{\{\textbf{definition}: any of several families of fishes…, \textbf{examples}: [], \textbf{hypernyms}: [spiny-finned fish, acanthopterygian], \textbf{hyponyms}: [flounder, halibut]\}} \\ \midrule
\multirow{2}{*}{\textbf{AWA2}} &
  \multirow{2}{*}{\textbf{Killer whale}} &
  \underline{\textbf{Text:}} Orcas (killer whales) are one of 35 species in the oceanic dolphin family, Delphinidae. \\ \cdashlinelr{3-3}
 &
   &
  \underline{\textbf{Structured:}} \textit{\{\textbf{appendages}: [flippers, tail], \textbf{behavior}: [fierce, smart, group], \textbf{color}: [black, white], \textbf{diet}: [fish, meat, plankton, hunter], \textbf{habitat}: [arctic, coastal, ocean, water], \textbf{mobility}: [swims, fast, strong, active, agility], \textbf{shape}: [big, bulbous, lean], \textbf{skin}: [patches, spots, hairless, toughskin]\}} \\   
  \bottomrule
\end{tabular}
}
\caption{
Examples of natural language and structured format descriptions for sample classes from each dataset.
Multiple descriptions are collected for each class, and we show a single one for conciseness.
}
\label{tab:example_descriptions}
\end{table*}

Example descriptions for both natural language and structured JSON descriptions for all four datasets are presented in Table \ref{tab:example_descriptions}, and description count and length statistics are presented in Table \ref{tab:descr_stats}.
\begin{table}[h]
\centering
\begin{tabular}{lcc}
\toprule
\textbf{Dataset} & \textbf{Num Descriptions} & \textbf{Description Lengths} \\ \midrule
RCV-1 & $17.9\pm 4.3$ & $17.4\pm 9.7$\\
20 NG & $19.2\pm 3.8$ & $17.7\pm 7.1$\\
CIFAR-100 & $20.3\pm 5.9$ & $16.7\pm 7.1$\\
AWA2 & $1250 \pm 0.0$ & $30.6 \pm 5.9$\\
\bottomrule
\end{tabular}
\caption{Statistics of the collected class descriptions including mean number of descriptions per class and mean lengths per description. Note that on AWA2 we automatically augment the descriptions, so there is no variance in the number of descriptions between classes.}
\label{tab:descr_stats}
\end{table}

\subsection{The effect of description quality}
\label{app:description_quality}
We assess the effect of train-time description quality on model performance by training \semsup{} with $n=10$ on CIFAR-100 and AWA2 with raw scraper descriptions without manual filtering. Performance on test across scenarios drop $<1$ point and is not statistically significant.
For example, the performance drops only from $61.0 \pm 1.7$ to $60.2 \pm 2.3$ for CIFAR-100 on the unseen classes scenario (S2).
These results suggest we can collect descriptions \textit{completely automatically} without any manual intervention.

\subsection{Collecting and Processing Descriptions}
\label{app:output_supervision}
The results were scraped from \url{www.google.com} and \url{www.duckduckgo.com} using a third-party scraping tool\footnote{\url{www.webscraper.io}}. Data collection was conducted between September 2021 and January 2022. To reduce variability, personalized results were turned off and regions were fixed to United States. Safe search was enabled for \url{www.google.com} and set to moderate on \url{www.duckduckgo.com}. The number of search returns for \url{www.google.com} was varied between 10 and 50. While we obtained more descriptions using a higher number of search returns, we found that the quality and relevance was often lower. 


\begin{figure}[h]
    \centering
    \includegraphics[width=0.4\linewidth]{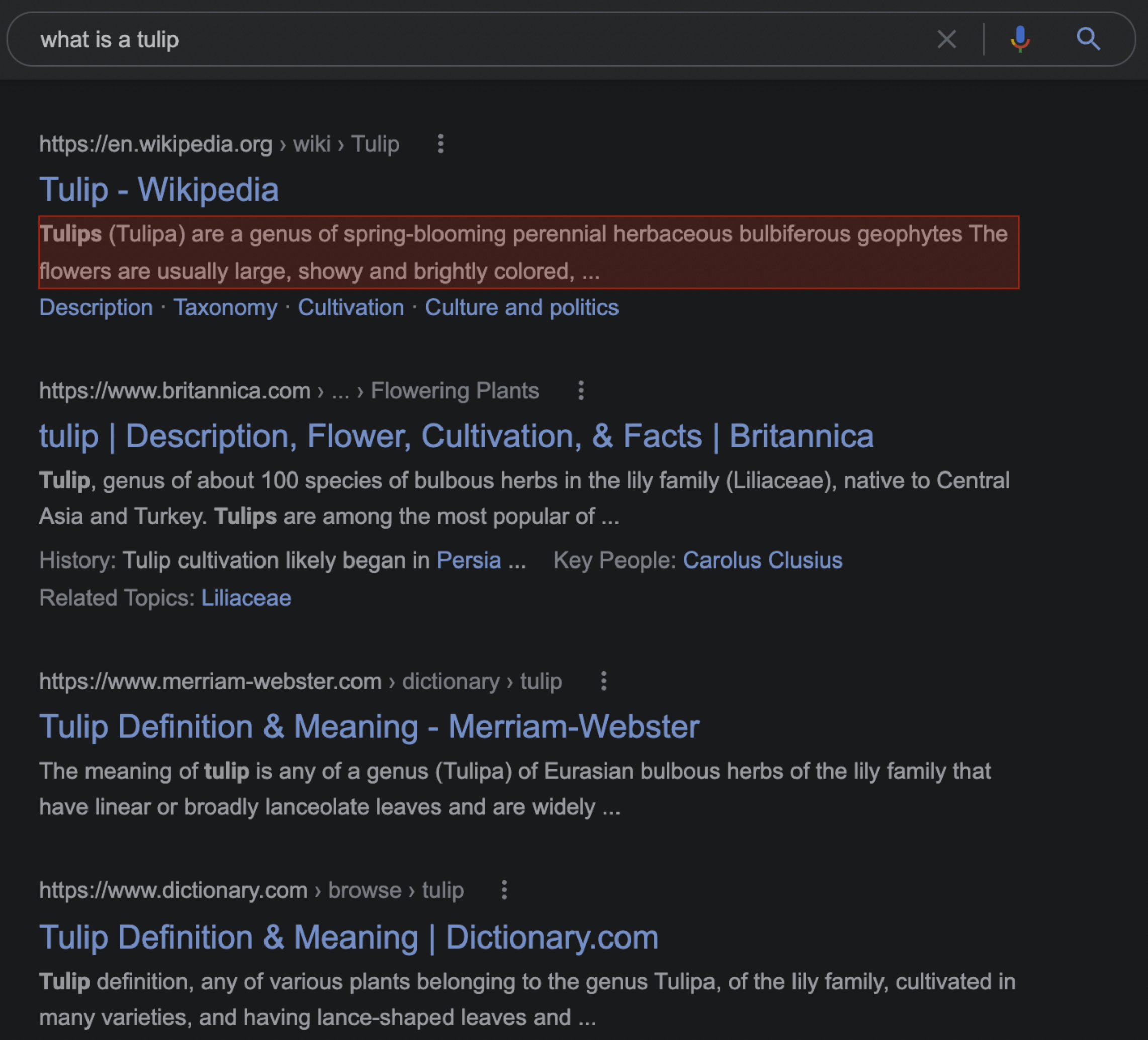}
    \caption{Example scraper selection (highlighted in orange) from the search query for the class \textit{tulip} in CIFAR100.}
    \label{fig:scraper}
\end{figure}

An example scraping target is presented in Figure \ref{fig:scraper}. We automatically filter the scraped preview blocks by removing any incomplete sentences. For multi-sentence descriptions, we only take the first sentence. Sentences that are less than 5 words are discarded. After automatic filtering, we manually inspect the descirptions and remove irrelevant descriptions. The mean number and lengths of the collected descriptions is presented in Table \ref{tab:descr_stats}.
On all datasets, we divide the class descriptions into a 60-20-20 train-val-test split.

To construct JSONs for 20 NG, we use the class hierarchy and class name to fill in the keys \textit{topic} and \textit{class}. We then use the first 3 relevant links on the Wikipedia article for each class as the field for \textit{tags} and 3 relevant links (if available) in the \textit{See Also} section for the field \textit{related}. We augment the JSONs using the same procedure as that for AWA2. See table \ref{tab:example_descriptions} for an example.

\paragraph{Collecting JSONs for RCV1}
We consider two attributes for RCV1, the definition, and related terms (example in Table~\ref{tab:example_descriptions}).
We get the definition using a similar procedure as above, by scraping the web using a query containing the class name.
To get the related term, we query the website \url{https://www.powerthesaurus.org} using the class name and retrieve the top-$20$ related terms (or fewer if there are lesser terms), and randomly sample $6$ terms to be in a description.
We construct $5$ such descriptions for each class.
We randomly chose these hyperparameters, and our procedure seems robust to their values.

\section{Datasets}
\label{sec:dataset}

Detailed dataset statistics including modality, size, and train-test split for classes and superclasses are presented in Table \ref{tab:dataset_stats}. 
\begin{table}[h]
\centering
\resizebox{0.5\linewidth}{!}{
\begin{tabular}{lcccc}
\toprule
\textbf{Dataset}    &  \textbf{20NG} & \textbf{RCV1} & \textbf{CIFAR100}  & \textbf{AWA2}\\ \midrule
Modality            & Text & Text & Image & Image\\
Size                & 20k & 800k & 50k & 37k\\
\# Classes (\#Test) & 20 (4) & 103 (25) & 100 (10) & 50 (10)\\
\# Supercls (\#Test)  & 5 (3) & 86 (17) & 20 (10) & NA \\
\bottomrule
\end{tabular}
}
\caption{Dataset statistics. \# Classes and \# Supercls indicate the number of classes and superclasses in total. \#Test in brackets indicates the number of those classes we reserve for testing.}
\label{tab:dataset_stats}
\end{table}

\subsection{RCV1}

RCV1 contains $800,000$ articles and we create a 60:20:20 split for train, validation, and test respectively.
It contains $103$ classes.

\subsection{20NG}
\begin{table}
\small
\centering
\begin{tabular}{@{}ll@{}}
\toprule
Val Classes & alt.atheism \\
& comp.sys.mac.hardware \\
& rec.motorcycles \\
& sci.electronics, \\
\midrule
Test Classes & comp.os.ms-windows.misc\\
& rec.sport.hockey\\
& sci.space\\
& talk.politics.guns\\
\midrule
Val Superclasses & recreation \\
& religion \\
\midrule
Test Superclasses & computer\\
& science\\
& politics\\
\bottomrule
\end{tabular}
\caption{Details for the 20 NG dataset. Training classes are the remaining 12 classes not in val classes or test classes.}
\label{tab:ng_info}
\end{table}
We use the 18828 variant for each newsgroup. Since the original dataset does not define train-test splits, we construct our own 80-20 train test split. We further divide the training set into training and validation sets with a porportion of 80-20.

We present details of the 20 NG dataset splits in table \ref{tab:ng_info}. When evaluating generalization to superclasses on 20 NG we remove the {\tt misc.forsale} class since it is its own superclass.

\subsection{CIFAR-100}. 
\begin{table}
\small
\centering
\begin{tabular}{@{}ll@{}}
\toprule
Val Classes & streetcar, rabbit, man\\
& lamp, forest, otter \\
& crab, crocodile, house \\
& orchid\\
\midrule
Test Classes & motorcycle, pine\_tree, bottle\\
& trout, chair, butterfly\\
& chimpanzee, orange, leopard\\
& possum\\
\midrule
Val Superclasses & large\_omnivores\_and\_herbivores\\
& medium\_mammals, people\\
& large\_man-made\_outdoor\_things\\
& insects, household\_electrical\_devices\\
& food\_containers, fish\\
& flowers, vehicles\_2\\
\bottomrule
\end{tabular}
\caption{Details for CIFAR-100. Training classes are the remaining 80 classes not in val classes or test classes. The test superclasses are the remaining 10 superclasses not listed in the val superclasses above.}
\label{tab:cifar_info}
\end{table}
We use the provided train-test split, but divide the train set 80-20 into training and validation examples.

\subsection{AWA2}
We use the predefined train-val-test splits of classes provided in the paper \cite{xian2018goodbadugly}. We use only the second of the three train-val splits provided. We split the instances into train and test examples 80-20 and further divide the training set 80-20 into training and validation examples.

To construct the JSON, we first assign each attribute to a parent attribute. The final class-level JSON consists of the parent attributes as keys, and the values are attributes that are present in the class. We augment this dataset by first adding 50 samples per class of corrupted examples, by randomly deleting attributes independently with probability 0.15, and then further multiplying this by 25 permutations.

\section{Model Training and Evaluation}
\label{app:training}
All models are end-to-end differentiable and we train them using the AdamW optimizer \cite{loshchilov2017AdamW}. We use a constant learning rate of $1\times10^{-4}$ for all the vision experiments on AWA2 and CIFAR-100 and a constant learning rate of $2\times10^{-5}$ for all experiments on 20 NG. For efficiency, the class descriptions are encoded into the output matrix $\semout_b$ at each minibatch, so that all instances in the batch share the same output matrix. We use the validation set for early stopping, and test checkpoints saved at the point of highest validation accuracy. All implementation was done in PyTorch and PyTorch Lightning and experiments were run on either a single NVIDIA RTX2080 or a single NVIDIA RTX3090.

\subsection{Using T5 for text datasets}
\label{app:t5_and_clip}
We use a T5-large~\cite{raffel2020exploring} fine-tuned on MNLI~\cite{MNLI} and check if a label description entails the input instance.
We use the probability of the \texttt{entailment} label as the compatibility between the input instance and the corresponding label description.
We repeat this for all the classes, and pick the class with the highest probability of entailment.
Since we need to run a forward pass for each instance-class pair (because of GPU memory limitations), we run the model on all the 20NG data but only $10\%$ of the RCV1 data.
We provide the script for this evaluation.

\section{Limitations and Risks}
\label{app:limitations}
There were no third-party human subjects involved in this study. We list a few limitations and risks of our work and how they can be mitigated:
\begin{enumerate}
    \item \semsup{} which uses natural language descriptions relies on scraping descriptions from search engines. This method worked even for RCV1 which contains niche classes. But it is possible that there are classification tasks which contains very domain specific classes for which scraping from the search engine does not provide good performance. In this case, we believe that we can use a small number of expert descriptions to provide a semantic understanding of the class to the model. We plan to run a study to judge the performance when crowd-sourced descriptions are used.
    \item While~\semsup{} significantly beats baselines and generalizes better to unseen scenarios, it is hard to pin point when generalization will work and when it will not. For example, if we trained on \textit{cat} and \textit{dog}, will the model generalize to \textit{wolf}? We believe that there is scope for better theoretical understanding of~\semsup{} which will allow users to reason about when \semsup{} can generalize and when it cannot.
    \item \semsup{} allows users to flexibly specify any class so long as a description is provided. However, if the class is very different from those seen during training, the model may provide wrong or misleading classifications. Future work that provide additional guarantees and indications for when zero-shot learning systems will fail will be helpful in mitigating these risks. 
\end{enumerate}

\end{document}